\documentclass[final]{cvpr}

\usepackage{times}
\usepackage{epsfig}
\usepackage{graphicx}
\usepackage{amsmath}
\usepackage{amssymb}
\usepackage{times}
\usepackage{epsfig}
\usepackage{graphicx}
\usepackage{amsmath}
\usepackage{amssymb}
\usepackage{booktabs}
\usepackage{multirow}
\usepackage{color}
\usepackage{commath}
\usepackage{xspace}
\usepackage{colortbl}
\usepackage{bm}
\usepackage{makecell}
\usepackage{soul}
\definecolor{citecolor}{RGB}{34,139,34}
\definecolor{grayDark}{gray}{0.95}
\definecolor{grayLight}{gray}{0.98}
\definecolor{darkgreen}{rgb}{0.8, 0.1, 0.1}

\newcommand{\et}{\emph{et al.}}

\newcommand{\myparagraph}[1]{{ \noindent \bf #1}}

\newcommand{\shape}{\mbox{\boldmath $\beta$}}
\newcommand{\pose}{\mbox{\boldmath $\theta$}}

\newcommand{\cam}{\mbox{\boldmath $\pi$}}

\usepackage[pagebackref=true,breaklinks=true,letterpaper=true,colorlinks,
  citecolor=citecolor,bookmarks=false]{hyperref}

\def\cvprPaperID{901} 
\def\confYear{CVPR 2021}

\begin{document}

\title{Body Meshes as Points}

\author{
Jianfeng~Zhang$^1$ \qquad
Dongdong~Yu$^2$    \qquad
Jun~Hao~Liew$^1$   \qquad
Xuecheng~Nie$^3$   \qquad
Jiashi~Feng$^1$  \\
\small{$^{1}$National University of Singapore} \qquad
\small{$^{2}$ByteDance AI Lab} \qquad
\small{$^{3}$Yitu Technology} \\
\small \texttt{\{zhangjianfeng,liewjunhao\}@u.nus.edu} \quad \texttt{yudongdong@bytedance.com}
\quad \texttt{elefjia@nus.edu.sg}
}

\maketitle

\begin{abstract}

We consider the challenging multi-person 3D  body mesh estimation task in this work. Existing methods are mostly two-stage based\textemdash one stage for person localization and the other stage for individual  body mesh estimation, leading to redundant pipelines with high computation cost and degraded performance for complex scenes (\eg,  occluded person instances).   
In this work, we present a single-stage model, Body Meshes as Points (BMP), to simplify the pipeline and lift both efficiency and performance. 
In particular, BMP adopts a new method that represents multiple  person instances as points in the spatial-depth space where each point is associated with one body mesh.
Hinging on such representations,  
BMP can directly predict body meshes for multiple persons in a single stage by concurrently localizing person instance points and estimating the corresponding body meshes.
To better reason about depth ordering of all the persons within the same scene, BMP designs a simple yet effective inter-instance ordinal depth loss to obtain depth-coherent  body mesh estimation. 
BMP also introduces a novel keypoint-aware augmentation to   enhance model robustness to occluded person instances. 
Comprehensive experiments on benchmarks Panoptic, MuPoTS-3D and 3DPW clearly demonstrate the state-of-the-art efficiency of BMP for multi-person body mesh estimation, together with outstanding accuracy. Code can be found at: \url{https://github.com/jfzhang95/BMP}.

\end{abstract}
\section{Introduction}

3D human body mesh recovery aims to reconstruct the 3D full-body mesh of the person instance from images or videos.
As a fundamental yet challenging task, it has been widely applied for action recognition~\cite{varol2019synthetic}, virtual try-on~\cite{mir2020learning}, motion retargeting~\cite{liu2019liquid}, \etc. 
With the recent notable progress in single-person based full-body mesh recovery~\cite{hmrKanazawa17,kolotouros2019convolutional,arnab2019exploiting,kocabas2020vibe}, a  more realistic and challenging setting has attracted increasing attention, \ie to estimate  body meshes for multiple persons from a single image.

\begin{figure}[t]
	\centering
	\includegraphics[width=0.9\linewidth]{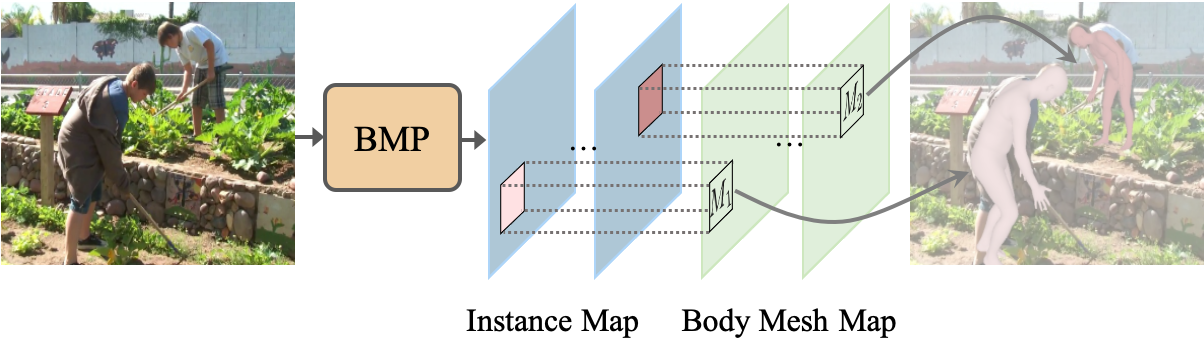}
	\caption{
	\textbf{Our single-stage solution.}
    The proposed model represents each person instance as the center point of its body. Instance localization and body mesh recovery are then directly predicted from the center point features, enabling simultaneous reconstruction of multiple persons in a single stage.
    }
	\label{fig:fig1}
\end{figure}

Existing methods for multi-person mesh recovery are mainly two-stage solutions, including  top-down~\cite{jiang2020coherent} and bottom-up~\cite{zanfir2018monocular} approaches.
The former first localizes person instances via a person detector, based on which it then recovers the 3D meshes individually;
the bottom-up approach estimates person keypoints at first, and then jointly reconstructs multiple 3D human bodies in the image via constrained optimization~\cite{zanfir2018monocular}.
Though with notable accuracy, the above paradigms are inefficient with computational redundancy. 
For instance, the former one estimates body mesh for each person separately, and consequently the total computation cost linearly grows with the number of persons in the image, while the latter requires grouping the keypoints into corresponding persons and inferring the body meshes iteratively, leading to high computational cost.

Targeted at a more efficient and compact pipeline, we consider exploring a single-stage solution.
Despite the recent popularity and promising performance of single-stage methods on 2D keypoints estimation~\cite{nie2019spm} and object detection tasks~\cite{zhou2019objects,tian2019fcos}, a single-stage pipeline for multi-person mesh recovery is barely explored as it remains unclear how to effectively integrate both person localization and mesh recovery steps within a single stage.
In this work, we propose a new instance representation for multi-person body mesh recovery that represents multiple person instances  as points in the spatial-depth space where each point is associated with one body mesh.  
Such an representation allows effective parallelism of person localization and body mesh recovery.
Based on it, we develop a new model architecture that exploits shareable features for both localization and mesh recovery and thus achieve a single-stage solution.

In particular, the model has two parallel branches, one for instance localization and the other for body mesh recovery.
In the localization branch, we model each person instance as a single point in a 3-dimensional space, \ie spatial (2D) and depth (1D),
where each localized point (detected person) is associated with a body mesh in the body mesh branch represented by the SMPL parametric model~\cite{loper2015smpl}.
This in turn converts the multi-person mesh recovery into a single-shot regression problem (Fig.~\ref{fig:fig1}). 
Specifically, the spatial location is represented by discrete coordinates \wrt regular grids over the input image.
Similarly, we discretize depth into several levels to obtain the depth representation. 
To learn better feature representation to differentiate instances at different depth, motivated by the phenomenon that a person closer to the camera tends to seem larger in the image, we adopt the feature pyramid network (FPN)~\cite{lin2017feature} to extract multi-scale features and use features from the lower scales to represent the closer (and larger) instances.
In this way, each instance is represented as one point, whose associated features (extracted from its corresponding spatial location and FPN scale) are used to effectively estimate its body mesh.
We name this 
\textbf{B}ody \textbf{M}eshes as \textbf{P}oints (\textbf{BMP}).

Applying the BMP model for estimating multi-person body mesh simultaneously faces two challenges in realistic scenarios: how to coherently reconstruct instances with correct depth ordering, and  how to handle the common occlusion issue (\eg, overlapping instances and partial observations).
For the first challenge, we consider explicitly using the ordinal relations among all the persons in the scene to supervise the model to learn to output body meshes with correct depth order.
However, obtaining such ordinal relations is non-trivial for the scenes captured in the wild, since there is no 3D annotation available. 
Inspired by the recent success of depth estimation for human body joints~\cite{moon2019camera,zhen2020smap}, we propose to take the depth of each person (center point) predicted by a model pre-trained on 3D datasets with depth annotations as the pseudo ordinal relation for model training on the in-the-wild data, which is experimentally proved beneficial to depth-coherent body mesh reconstruction.

Also, to tackle the common occlusion issue, we propose a novel keypoint-aware occlusion augmentation strategy to improve the model robustness to occluded person instances.
Different from the previous method~\cite{sarandi2018robust} that randomly simulates occlusion in images, we generate synthetic occlusion based on the position of skeleton keypoints.
Such keypoint-aware occlusion explicitly forces the model to focus on body structure, making it more robust to occlusion.

Comprehensive experiments on 3D pose benchmarks Panoptic~\cite{joo2015panoptic}, MuPoTS-3D~\cite{singleshotmultiperson2018} and 3DPW~\cite{vonMarcard2018} evidently demonstrate the high efficiency of the proposed model.
Moreover, it achieves new state-of-the-art on Panoptic and MuPoTS-3D datasets, and competitive performance on 3DPW dataset. 
Our contributions are summarized as follows:
1) To our best knowledge, we are among the first to explore the \textit{single-stage} solution to multi-person mesh recovery. We introduce a new person instance representation that enables simultaneous person localization and body mesh recovery for all person instances in an image within a single stage, and design a novel model architecture accordingly.
2) We propose a simple yet effective inter-instance ordinal relation supervision to encourage depth-coherent reconstruction.
3) We propose a keypoint-aware occlusion augmentation strategy that takes body structure into consideration, to improve model robustness to occlusion.

\begin{figure*}[h!]
	\centering
	\includegraphics[width=0.95\linewidth]{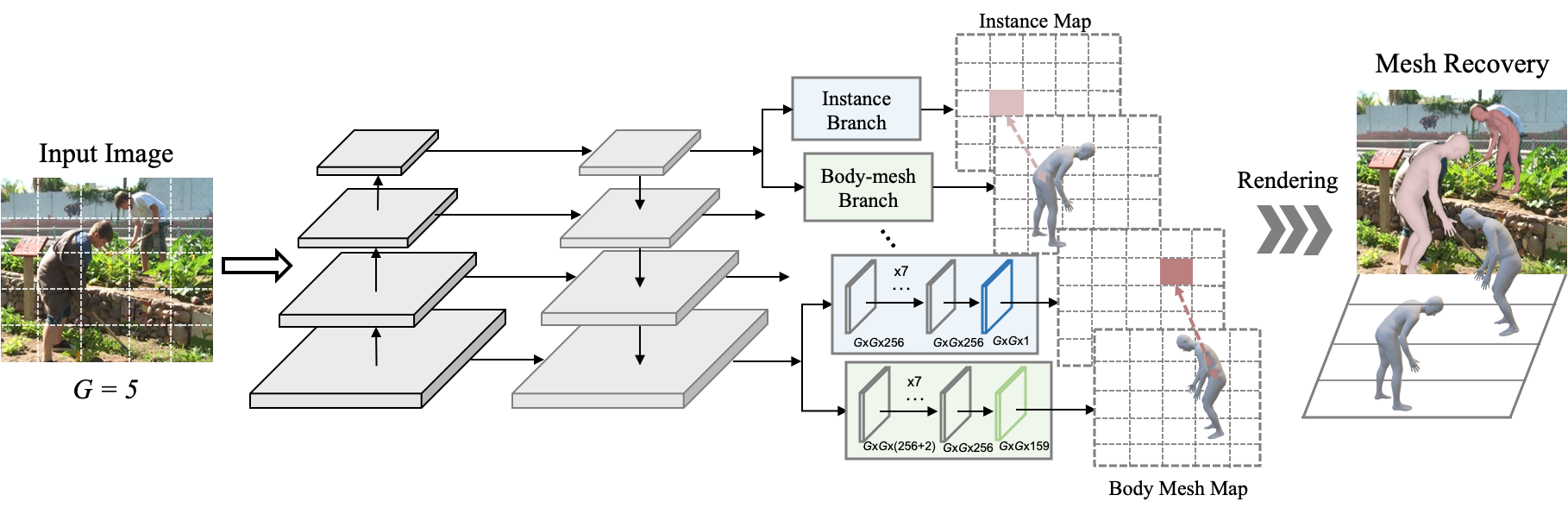}
	\caption{\textbf{Illustration of our BMP framework.} 
     An input image is uniformly divided into $G\times G$ grids with $G = 5$ in this example. The model adopts an FPN with $K$ levels ($K=4$ here). Each person instance is thus   represented by its residing grid cell and its associated FPN level (according to its depth). BMP uses the features from the grid cell and FPN level to localize the contained person (top) and estimate the body mesh (bottom) simultaneously.
    }
    \label{fig:framework}
\end{figure*}
\section{Related Work}

\myparagraph{Single-person 3D pose and shape}
Previous works estimate 3D poses in the form of body skeleton~\cite{martinez2017simple,mehta2017vnect,tome2017lifting,zhou2017towards,popa2017deep,pavlakos2018ordinal,sun2018integral,zhang2020inference,gong2021poseaug} or non-parametric 3D shape~\cite{gabeur2019moulding,smith2019facsimile,varol2018bodynet}. 
In this work, we use the 3D mesh to represent the full-body pose and shape, and adopt the SMPL parametric model~\cite{loper2015smpl} for body mesh recovery. 
In literature, Bogo~\et~\cite{bogo2016keep} proposed SMPLify, the first optimization-based method to fit SMPL on the detected 2D joints iteratively. 
Later works extend SMPLify by either using more dense reference points to replace sparse keypoints like silhouettes and voxel occupancy grids for SMPL fitting~\cite{lassner2017unite,varol2018bodynet}, or fitting a more expressive model (\eg, SMPL-X) than SMPL~\cite{pavlakos2019expressive}.

Some recent works directly regress the SMPL parameters from images via deep neural networks in a two-stage manner. 
They first estimate the intermediate representation (\eg, keypoints, silhouettes, etc) from images and then map it to SMPL parameters~\cite{pavlakos2018learning,omran2018neural,tung2017self,kolotouros2019convolutional}. 
Some others directly estimate SMPL parameters from images, either using complex model training strategies~\cite{hmrKanazawa17,guler2019holopose} or leveraging temporal information~\cite{arnab2019exploiting,kocabas2020vibe}.
Although high accuracy is achieved in single-person cases,  it remains unclear how to extend them to the more general multi-person cases.

\myparagraph{Multi-person 3D pose and shape}
For multi-person 3D pose estimation, most existing methods adopt a top-down paradigm~\cite{rogez2017lcr,dabral2018learning,rogez2019lcr}.
They first detect each person instance and then  regress the locations of the body joints.
Follow-up improvements are made by estimating additional absolute depth~\cite{moon2019camera}, considering multi-person interaction~\cite{guo2020pi,li2020hmor} or extending to whole-body pose estimation~\cite{weinzaepfel2020dope}.
Alternatively, some approaches also explore the bottom-up paradigm. 
SSMP3D~\cite{mehta2018single} and SMAP~\cite{zhen2020smap} estimate 3D poses from occlusion-aware pose maps and use Part Affinity Fields~\cite{cao2017realtime} to infer their association.
LoCO~\cite{fabbri2020compressed} maps the image to the volumetric heatmaps and then estimates multi-person 3D poses from them by an encoder-decoder framework.
PandaNet~\cite{benzine2020pandanet} is an anchor-based model where 3D poses are regressed for each anchor position.

In contrast to the prosperity of multi-person 3D pose estimation, there is a limited number of works denoted to  body mesh recovery for multiple people.
Zanfir~\et~\cite{zanfir2018monocular} first estimate 3D joints of persons in the image and then optimize their 3D shapes jointly with multiple constraints.
They also propose a two-stage regression-based scheme that first estimates 3D joints for all the persons and then regresses their 3D shapes based on these 3D joints~\cite{zanfir2018deep}.
Instead of regressing SMPL parameters from an intermediate representation (\eg, 3D joints), Jiang~\et~\cite{jiang2020coherent} attach an SMPL head to the Faster R-CNN framework~\cite{ren2015faster} for estimating SMPL parameters directly from the input image in a top-down manner.
Despite the encouraging results, these methods are based on the indirect multi-stage framework and  suffer low efficiency. 
Different from all previous methods that rely on a multi-stage pipeline with computation redundancy, our method unifies person localization and body mesh, and enables a box-free and (ad hoc) optimization-free single stage solution to multi-person body mesh recovery. 

\myparagraph{Point-based representation}
The point-based methods~\cite{duan2019centernet,zhou2019objects,tian2019fcos} represent instances by a single point at their center. 
This approach is regarded as a simple replacement of the anchor-based representation, which has been widely used in many tasks, including object detection~\cite{duan2019centernet,zhou2019objects,tian2019fcos}, 2D keypoints estimation~\cite{nie2019spm} and instance segmentation~\cite{wang2019solo}.
However, these methods cannot be directly applied to body mesh recovery. 
In this work, we extend the point-based representation to multi-person body mesh recovery. 
A concurrent work~\cite{CenterHMR} adopts a similar solution to body mesh recovery. Our model differs from it in two significant aspects: 
1)  BMP aims at more coherent reconstruction of persons in the scenes. It handles challenging spatial arrangement and occlusion problems by exploiting the
ordinal depth loss and the keypoint-aware augmentation strategy, which are not considered in \cite{CenterHMR}. 2) BMP
adopts a novel 3D point-based representation to differentiate instances at different depths, thus is more robust to overlapped instances; whereas \cite{CenterHMR} uses only 2D representation, and would fail in such cases.

\section{Body Meshes as Points}

\subsection{Proposed single-stage solution} \label{sec:formulation}

Given an image $I$, multi-person body mesh recovery targets at recovering body meshes of all the person instances in $I$. 
Existing approaches~\cite{zanfir2018deep,zanfir2018monocular,jiang2020coherent} solve this task via sequentially localizing and estimating the body mesh in a multi-stage manner, leading to computation redundancy. 
Differently, this work aims to unify the instance localization and body mesh recovery into a single-stage solution to enable a more efficient and concise framework.

We represent each person instance as a single point $(i,j,k)$ in a 3-dimensional space (spanned by 2D spatial and 1D depth dimensions).
By dividing the input image uniformly into $G\times G$ grids, its spatial dimension can be easily represented within such a grid coordinate.  
If the body center of a person falls into grid cell $(i,j)$, it is assigned with spatial coordinate $(i,j)$. 
Similarly, for the depth dimension, we discretize the depth value to $K$ levels and obtain the $k$ value for each instance according to its depth. 
Such discretized depth value is beneficial for handling occluding instances, especially when the body centers of multiple instances fall into the same spatial grid coordinate.

Given this representation, we reformulate multi-person mesh recovery as two simultaneous prediction tasks: 1) instance localization and 2) body mesh recovery.

\myparagraph{Instance localization}
For the first task, we employ the \textit{instance map} $\boldsymbol{\mathcal{C}}=\{\boldsymbol{\mathcal{C}}_1,\ldots,\boldsymbol{\mathcal{C}}_K\}$, where $\boldsymbol{\mathcal{C}}_k \in \mathbb{R}^{G\times G\times 1}$, to locate each person instance in the image, where $G$ denotes the number of grid cells along one side, while $K$ refers to the number of total depth levels.
For each depth level, the network is trained to regress a scalar indicating the probability of every grid cell containing a person.

To construct ground truth (GT) for training, we
first determine the depth value $k$ for each instance.
We observe that a person tends to seem larger (smaller) in the image when standing closer to (away from) the camera. 
In other words, the depth of an instance is roughly inversely proportional to its scale.
Inspired by it, we employ a Feature Pyramid Network (FPN)~\cite{lin2017feature} with $K$ pyramid levels to capture $K$ different scales, each of which is used to represent the instance with the corresponding depth. 
More specifically, for each instance, we compute its scale $s=\sqrt{hw}$ where $(h, w)$ denotes the GT body size, and associate it to the corresponding pyramid level $k$, according to Table \ref{tab:fpn_seg}.

\begin{table}[h!]
    \centering
    \scalebox{0.84}{
    \begin{tabular}{l|ccccc}
    \Xhline{1pt}
        Pyramid & P$_2$ & P$_3$ & P$_4$ & P$_5$ & P$_6$\\
        \hline
        Stride & 8 & 8 & 16 & 32 & 32\\
        \hline
        Grid number $G$ & 40 & 36 & 24 & 16 & 12\\
        \hline
        Instance scale $s$ &$<$64 & 32$\sim$128 & 64$\sim$256 & 128$\sim$512 & $\geq$256\\
    \Xhline{1pt}
    \end{tabular}}
    \caption{We employ FPN with five pyramid levels. P$_{k{+}1}$ is used to predict instance $\boldsymbol{\mathcal{C}}_k$ and body mesh maps $\boldsymbol{\mathcal{P}}_k$, where $k=1,\ldots ,5$. 
    } 
     \label{tab:fpn_seg}
     \vspace{-2mm}
\end{table}

Next, we locate the grid cell $(i, j)$ in $\boldsymbol{\mathcal{C}}_k$ where the \textit{central region} of that person lies. 
Inspired by \cite{zhou2019objects,duan2019centernet}, the central region is defined as follows:
given the GT body center $(x^c, y^c)$,  body size $(h, w)$ of each person and a controllable scale factor $\epsilon$, the position and size of the central region are defined as $(x^c, y^c, \epsilon w, \epsilon h)$. 
In this work, we set the position of pelvis as body center and $\epsilon=0.2$.
Once identified, the grid cell $(i, j)$ of the $k$-th pyramid level, \ie, $\boldsymbol{\mathcal{C}}_k(i,j)$ is labeled as positive (label 1).
The above steps are repeated for all the instances in the image.

\myparagraph{Body mesh representation}
In parallel with the instance localization, we use the \emph{body mesh map} $\boldsymbol{\mathcal{P}}=\{\boldsymbol{\mathcal{P}}_1,\ldots,\boldsymbol{\mathcal{P}}_K\}$, for body mesh recovery, where $\boldsymbol{\mathcal{P}}_k \in \mathbb{R}^{G\times G\times S}$ and $S$ is the dimension of body mesh representation.
Concretely, given a positive response in $\boldsymbol{\mathcal{C}}$ that indicates the presence of a person, we regress the body mesh representation using the features from the corresponding grid cell, as shown in Fig.~\ref{fig:framework}. 
In this work, we use the SMPL parametric model~\cite{loper2015smpl} for body mesh representation, which renders a body mesh using the   pose parameters $\pose \in \mathbb{R}^{72}$ and shape parameters  $\shape \in \mathbb{R}^{10}$. 
To improve the training stability, we adopt the 6D rotation representation~\cite{zhou2019continuity} for the pose parameters with $\pose \in \mathbb{R}^{144}$. 
The body mesh map also predicts a camera parameter $\cam=\{s, t_x, t_y\} \in \mathbb{R}^{3}$ for projecting body joints from 3D back to 2D space, which enables training on in-the-wild 2D pose datasets~\cite{johnson2010clustered,lin2014microsoft,andriluka20142d} to improve model generalization~\cite{hmrKanazawa17}.
We further introduce a scalar confidence score $c$ defined as the OKS~\cite{girdhar2018detect} between the projected and GT 2D keypoints, to reflect the confidence level of the SMPL prediction; and we also propose an absolute depth variable $d$ for the corresponding person instance that will be used for penalizing body mesh estimations with incoherent depth ordering (see  Sec.~\ref{sec:orloss} for details).  
Therefore, the total channel number of the body mesh map $S$ is 159.

\myparagraph{Network architecture}
We employ ResNet-50~\cite{He_2016_CVPR} as our backbone. FPN is built on top of the backbone to extract a pyramid of feature maps (256-d). 
To perform body mesh recovery, we attach two task-specific heads to each level of the feature pyramid, one for instance localization and the other for the corresponding body mesh recovery, 
responsible for obtaining the instance map $\boldsymbol{\mathcal{C}}_k \in \mathbb{R}^{G \times G \times 1}$ and the body mesh map $\boldsymbol{\mathcal{P}}_k \in \mathbb{R}^{G \times G \times 159}$, respectively.
As shown in Fig.~\ref{fig:framework}, each head consists of 7 stacked $3\times 3$ convolutions and one task-specific prediction layer.
However, directly estimating the camera parameter from the whole image is non-trivial since it is sensitive to instance position.
Inspired by CoordConv~\cite{liu2018intriguing}, we concatenate normalized pixel coordinates to the input feature map at the beginning of the mesh recovery head to encode position information into the network for better estimating camera parameter.
Additionally, Group Normalization~\cite{wu2018group} is used in both prediction heads for facilitating model training. 
In order to match the features of size $H \times W$ to $G \times G$, we apply bilinear interpolation before the instance and body-mesh recovery branch.

\subsection{Inter-instance ordinal depth supervision} \label{sec:orloss}
Multi-person body mesh recovery is inherently ill-posed as multiple 3D predictions can correspond to the same 2D projection.
Therefore, the trained model would produce ambiguous body mesh estimations with incorrect depth order due to lack of priors.
To alleviate such a problem, we use ordinal depth relations among all the persons in the input as supervision to guide reasoning about the depth ordering during the training process.

More concretely, given any two persons ($p_m$, $p_n$) in the image, we define the ordinal depth relation between them as $R(p_m, p_n)$, taking the value:
\begin{equation}
    R(p_m, p_n) = 
    \begin{cases}
        +1, & \text{ if } d_n-d_m>T, \\
        -1, & \text{ if } d_m-d_n>T, \\
        0, & \text{ if } |d_m-d_n|\leq T,
    \end{cases}
    \vspace{-2mm}
\end{equation}
where $d_m \in \mathbb{R}^{1}$ denotes the depth of the person $p_m$ and $T$ is a pre-defined threshold to determine the ordinal relation. 
The ordinal relation $R(p_m, p_n)=0$ means both instances are at roughly the same depth; otherwise one of them is closer to the camera than the other.
With the ordinal relation of ($p_m$, $p_n$), we define the ordinal depth loss for this pair as
\begin{equation}
    L_{(p_m, p_n)} = 
    \begin{cases}
        \log(1+\exp{(z_m-z_n))}, &\text{ if } R=+1, \\
        \log(1+\exp{(z_n-z_m))}, &\text{ if } R=-1, \\
        (z_m-z_n)^2, &\text{ if } R=0,
    \end{cases}
\end{equation}
where $z_m=\frac{2f}{s_m\alpha}$ denotes $m$-th person's body mesh depth calculated from the predicted camera parameter with focal length $f$, scale $s_m$ and images's long edge width $\alpha$.
The ordinal depth loss enforces a large margin between $z_m$ and $z_n$ if $R(p_m, p_n)\neq 0$, \ie, one of them measured as closer than the other, and otherwise enforces them to be equal.

However in practice, such ordinal depth relations are rarely available for the scenes captured in the wild due to lack of 3D annotations. 
To solve this issue, we propose to use pseudo ordinal relations for model training on the in-the-wild data.
Specifically, we first train the model on 3D datasets~\cite{ionescu2014human3,singleshotmultiperson2018} with depth annotations to learn to estimate the depth of each person in the images.
We define the depth $d$ of each person as the depth of body center (\ie, pelvis joint). 
The model is trained by minimizing a depth loss $L_{depth}$, which is defined as the mean square errors (MSE) between the predicted and GT depths. 
After that, given an unlabeled data, we first leverage the pre-trained model to estimate the depth which is then used to obtain the pseudo ordinal relations for all the people in the image. Finally, given the pseudo ordinal relations, we adopt an OKS score-weighted ordinal depth loss to supervise the model training for images in the wild. The total loss for image $I$ is computed as the average loss of all instances pairs:
\begin{equation} \label{eqn:rank}
    L_{rank}=\frac{1}{N}\sum_{(p_m,p_n)} c_mc_nL_{(p_m, p_n)},
    \vspace{-2mm}
\end{equation}
where $N$ denotes the number of paired instances in the image, $c_m$ denotes OKS-score of the $m$-th person.
Intuitively, training the model with such inter-instance ordinal depth supervision can help the model build a global understanding of the depth layout in the input scene and thus ensure more coherent reconstructions.

\vspace{-1mm}
\subsection{Keypoint-aware occlusion augmentation} \label{sec:occaug}
SMPL-based body mesh recovery is highly sensitive to (partial) occlusion (\eg, overlapping persons, truncation)~\cite{zhang2020object,rockwell2020full}.
To improve model robustness to occlusion without requiring extra training data and annotations, we propose a keypoint-aware occlusion augmentation strategy during the training process.
The proposed augmentation strategy aims to generate synthetic occlusion to synthesize real challenging cases such as partial observation for model training.
Compared with previous work~\cite{sarandi2018robust} that randomly simulates occlusion on the images, which may produce easy training samples that are less helpful for boosting model performance,
our method directly generates synthetic occlusion based on the positions of skeleton keypoints, which can force the model to pay more attention to the body structure, leading to notable enhancement. 
More concretely, given a set of $J$ keypoints $\{\boldsymbol{j}_1, \ldots,\boldsymbol{j}_J\}$ of a person in the image, we first randomly choose a keypoint $\boldsymbol{j}_i$. 
Then we randomly sample a non-human object from the PASCAL VOC~\cite{everingham2011pascal} dataset and composite it at the location of the selected keypoint $\boldsymbol{j}_i$. 
We randomly resize the sampled object to the range of $[0.1*A, 0.2*A]$ before compositing, where $A=wh$ denotes the area of that person.
Additionally, we randomly shift the keypoint position by an offset $\boldsymbol{\delta}$ to avoid over-fitting.
During training, we set the probability of the occlusion augmentation as 0.5. 

\subsection{Training and inference} \label{sec:train_inference}

\myparagraph{Training}
For training our proposed BMP model, we define the loss function $L$ as follows:
\begin{equation}
L = L_{inst} + L_{mesh} + L_{depth} + 0.1*L_{rank},
\end{equation}
where $L_{inst}$ is a modified two-class Focal Loss~\cite{lin2017focal} for instance localization; $L_{depth}$ is the depth loss (Sec.~\ref{sec:orloss}); $L_{mesh}$ is the loss for body mesh estimation. 
The training details of the body-mesh branch are similar to those in HMR~\cite{hmrKanazawa17}. Specifically, we formulate $L_{mesh}$ as
\begin{equation} \label{eqn:mesh}
\begin{split}
    L_{mesh} &= L_{pose} + L_{vert} + \lambda_{3D}L_{3D} + \lambda_{2D}L_{2D} \\ &+ \lambda_{shape}L_{shape}+ \lambda_{conf}L_{conf} + \lambda_{adv}L_{adv}.
\end{split}
\vspace{-2mm}
\end{equation}
Here $L_{pose}$, $L_{shape}$, $L_{3D}$, $L_{vert}$ denote MSE between the predicted and GT pose and shape parameters as well as 3D keypoints and vertices, respectively. 
$L_{2D}$ is the 2D keypoints loss that minimizes the distance between the 2D projection from 3D keypoints and GT 2D keypoints.
$L_{conf}$ is the MSE of the predicted and GT confidences, where the GT confidence is computed as the OKS~\cite{girdhar2018detect} between the projected and GT 2D keypoints.  
Moreover, we use a discriminator and apply an adversarial loss $L_{adv}$ on the regressed pose and shape parameters, to encourage the outputs to lie on the distributions of real human bodies.
$\lambda_{3D}=4$, $\lambda_{2D}=4$, $\lambda_{shape}=0.01$, $\lambda_{conf}=1$ and $\lambda_{adv}=0.01$ are the weights of the corresponding loss terms.
The loss $L_{mesh}$ is applied independently to each positive grid cell. 
The ordinal depth loss $L_{rank}$ illustrated in Eqn.~(\ref{eqn:rank}) is adopted when the image contains more than one instance.

\myparagraph{Inference}
The overall inference procedure for BMP 
is illustrated in Fig.~\ref{fig:framework}. 
Given an image, BMP first obtains the instance map $\boldsymbol{\mathcal{C}}$ and the body mesh map $\boldsymbol{\mathcal{P}}$ from the prediction heads. 
Then it performs max pooling operation to find the local maximum on $\boldsymbol{\mathcal{C}}$ to obtain center point positions $\{ (x^c_i, y^c_i, k^c_i) \}_{i=1}^{\hat{N}}$, where $k^c_i$ and ($x^c_i, y^c_i$) denote the pyramid level and body center location for $i$-th person, respectively, and $\hat{N}$ is the number of estimated persons.
After that, BMP extracts the body mesh parameters of each person $p_i$ via $\boldsymbol{\mathcal{P}}_{k^c_i}(x^c_i, y^c_i)$. 
Finally, BMP outputs body mesh estimations by deforming the SMPL model using the predicted parameters. 
A keypoint-based NMS~\cite{girdhar2018detect} is applied to remove the redundant predictions if they exist.
We take the multiplication of the predicted OKS score and the probability score from the instance map as the confidence score for NMS.

\subsection{Implementation details} \label{sec:implementation}
We implement BMP with PyTorch~\cite{paszke2017automatic} and mmdetection library~\cite{mmdetection} and utilize Rectified Adam~\cite{liu2019variance} as the optimizer with an initial learning rate of $1e{-}4$.  
We resize all images to 832$\times$512 while keeping the same aspect ratio following the original COCO training scheme~\cite{su2019multi,wang2019solo,jiang2020coherent}.
During training, we augment the samples with horizontally flip and keypoints-aware occlusion (Sec.~\ref{sec:occaug}). 
Flip augmentation is conducted during testing. 
Moreover, since the BMP model directly extracts image-level features for estimations instead of features from cropped bounding boxes, it can take images with smaller resolution (512$\times$512) as inputs. 
We denote such a setting as \textit{BMP-Lite}. Other training and testing settings are the same between BMP-Lite and BMP. 
Please refer to supplementary for more details.

\section{Experiments}
In this section, we aim to answer following questions.
1) Can BMP provide both efficient and accurate 
multi-person mesh recovery?
2) Is BMP able to give coherent meshes for multiple persons with correct depth ordering?
3) Is BMP robust to cases where person instances are occluded or partially observed?
To this end, we conduct extensive experiments on several large-scale benchmarks.

\subsection{Datasets} \label{sec:datasets}

\myparagraph{Human3.6M}~\cite{ionescu2014human3} is the most widely used single-person 3D pose benchmark collected in an indoor environment.
It contains 3.6 million 3D poses and corresponding videos for 15 subjects. 
Due to its high-quality annotations, we use it following \cite{jiang2020coherent} for both training and testing.

\myparagraph{Panoptic}~\cite{joo2015panoptic} is a large-scale dataset captured in the Panoptic studio, offering 3D pose annotations for multiple people engaging in diverse social activities. 
We use this dataset for evaluation with the same protocol as~\cite{zanfir2018monocular}.

\myparagraph{MuPoTS-3D}~\cite{singleshotmultiperson2018} 
is a multi-person dataset with 3D pose annotations for both indoor and in-the-wild scenes.
We follow \cite{singleshotmultiperson2018} and use it for evaluation.

\myparagraph{3DPW}~\cite{vonMarcard2018} is a multi-person in-the-wild dataset, which features diverse motions and scenes. It contains 60 video sequences (24 train, 24 test, 12 validation) with full-body mesh annotations.
To verify generalizability of the proposed model to challenging in-the-wild scenarios, we use its test set for evaluation, following the same protocol as~\cite{kocabas2020vibe}.

\myparagraph{MPI-INF-3DHP}~\cite{mehta2017vnect} is a single-person multi-view 3D pose dataset. It contains 8 actors performing 8 activities, captured from 14 cameras. Mehta~\et,~\cite{singleshotmultiperson2018} generate a multi-person dataset called MuCo-3DHP, from MPI-INF-3DHP via mixing up segmented foreground human appearance. We use both datasets for training.

\noindent
\textbf{COCO}~\cite{lin2014microsoft}, \textbf{LSP}~\cite{johnson2010clustered}, \textbf{LSP Extended}~\cite{Johnson11}, \textbf{PoseTrack}~\cite{Andriluka_2018_CVPR}, \textbf{MPII}~\cite{andriluka20142d}
are in-the-wild datasets with annotations for 2D joints. 
We use them for training with the weakly-supervised training strategy~\cite{hmrKanazawa17} (Eqn.~(\ref{eqn:mesh})).

\subsection{Comparison with state-of-the-arts} \label{sec:sota}

\myparagraph{Single-person setting}
We first evaluate our proposed BMP model on the single-person setting to validate the strategy of BMP on factorizing the instance localization and mesh recovery does not sacrifice on the performance.
Concretely, we evaluate and compare the performance of BMP on the large-scale Human3.6M dataset with most competitive approaches~\cite{hmrKanazawa17,jiang2020coherent} sharing the similar regression target and learning strategy.
The results are shown in Table~\ref{tab:h36m}. We can observe BMP outperforms all these methods.

\begin{table}[!h]
\setlength\tabcolsep{2.5mm}
\footnotesize
\centering
	\begin{tabular}{c|c|c|c}
		\Xhline{1pt}
		Method & HMR~\cite{hmrKanazawa17} & CRMH~\cite{jiang2020coherent} & BMP  \\
		\hline 
		PA-MPJPE & 56.8 & 52.7 & 51.3 \\
		\Xhline{1pt}
	\end{tabular}
	\caption{\textbf{Results on Human3.6M.} We use mean per joint position errors in mm after Procrustes alignment (PA-MPJPE) as metric.
	}
	\label{tab:h36m}
	\vspace{-2mm}
\end{table}

\myparagraph{Multi-person settings}
Then we evaluate our BMP model for multi-person body mesh recovery. We first evaluate it on the multi-person dataset captured in the indoor Panoptic Studio~\cite{joo2015panoptic} and compare with the most competitive approaches~\cite{zanfir2018monocular,zanfir2018deep,jiang2020coherent}.
As shown in Table~\ref{tab:panoptic}, our BMP model achieves the best performance in all scenarios.
Overall, it improves upon the state-of-the-art top-down model CRMH~\cite{jiang2020coherent}  by 5.4\% (135.4 mm \vs 143.2 mm in MPJPE), while offering a faster inference speed\footnote{We count per-image inference time in seconds. For all methods, the time is counted on GPU Tesla P100 and CPU Intel E5-2650 v2 @ 2.60GHz, without using test-time augmentation.}. 
Moreover, it significantly outperforms CRMH for Ultimatum and Pizza scenarios with crowded scenes and severe occlusion, verifying its robustness to occlusion cases. 
In addition, its lite version,  BMP-Lite, is even faster, which only requires 0.038s to process an image, about $2\times$ faster than CRMH while achieving comparable performance.
These results demonstrate both 
effectiveness and efficiency of BMP for estimating body meshes of multiple persons in a single stage.

\begin{table}[!h]
\centering
\footnotesize
\setlength{\tabcolsep}{1.5mm}{
	\begin{tabular}{l|c|c|c|c|c|c}
	\Xhline{1pt}
	Method & Haggl. & Mafia & Ultim. & Pizza & Mean & Time[s] \\
	\hline
	Zanfir~\et~\cite{zanfir2018monocular} & 140.0 & 165.9 & 150.7 & {156.0} & 153.4 & - \\ 
	MubyNet~\cite{zanfir2018deep} & 141.4 & 152.3 & {145.0} & 162.5 & 150.3 & - \\ 	
	CRMH~\cite{jiang2020coherent} & 129.6 & {133.5} & 153.0 & 156.7 & {143.2} & 0.077 \\
	BMP-Lite & 124.2 & 138.1 & 155.2 & 157.3 & 143.7 & \textbf{0.038} \\
	BMP & \textbf{120.4} & \textbf{132.7} & \textbf{140.9} & \textbf{147.5} & \textbf{135.4} & 0.056 \\
	\Xhline{1pt}
	\end{tabular}}
	\caption{\textbf{Results on the Panoptic.} We use MPJPE as evaluation metric. The lower the better. Best in \textbf{bold}.}
	\label{tab:panoptic}
\end{table}

Another popular 3D pose estimation benchmark is the MuPoTS-3D dataset~\cite{mehta2017vnect}. We compare our method against two strong baselines, 
1) the combination of OpenPose~\cite{cao2018openpose} with single-person mesh recovery methods (SMPLify-X~\cite{pavlakos2019expressive} and HMR~\cite{hmrKanazawa17}),
and 2) the state-of-the-art top-down approach CRMH~\cite{jiang2020coherent}.
We report the results in Table~\ref{tab:mupots}. 
As we can see, BMP outperforms significantly previous methods on both evaluation protocols.

\begin{table}[!h]
\footnotesize
    \setlength\tabcolsep{4mm}
    \centering
	\begin{tabular}{l|c|c|c}
	\Xhline{1pt}
	Method & All & Matched & Time[s]\\
	\hline
	SMPLify-X~\cite{pavlakos2019expressive} & 62.84 & 68.04 & 6.4 \\
	HMR~\cite{hmrKanazawa17} & 66.09 & 70.90 & 0.26 \\	
	CRMH~\cite{jiang2020coherent} & {69.12} & {72.22} & 0.083 \\
	BMP-Lite & 68.63 & 71.92 & \textbf{0.038} \\
	BMP & \textbf{73.83} & \textbf{75.34} & 0.056 \\
	\Xhline{1pt}
	\end{tabular}
	\caption{\textbf{Results on MuPoTS-3D.} The numbers are 3DPCK. We report the overall accuracy (All), and the accuracy only for person annotations matched to a prediction (Matched). Best in \textbf{bold}.}
	\label{tab:mupots}
\end{table}

Lastly, we compare our BMP model with state-of-the-art approaches on the challenging in-the-wild 3DPW dataset.
Some approaches use the self-training strategy (\ie, SPIN~\cite{kolotouros2019spin}) or  temporal information (\ie, VIBE~\cite{kocabas2020vibe}), and they rely on off-the-shelf person detectors~\cite{cao2018openpose,redmon2018yolov3}. 
As shown in Table~\ref{tab:3dpw}, our BMP outperforms CRMH~\cite{jiang2020coherent} and SPIN~\cite{kolotouros2019spin} in terms of 3DPCK while maintaining an attractive efficiency, and achieves comparable results with VIBE~\cite{kocabas2020vibe} without relying on any temporal information.
Additionally, BMP-Lite obtains roughly the same performance as the state-of-the-art CRMH model while achieving $2.1\times$ faster inference speed. 
There results further confirm the effectiveness of our single-stage solution over existing multi-stage strategies, with very competitive efficiency.

\begin{table}[!h]
\setlength\tabcolsep{4.5pt}
\centering
\footnotesize
    \begin{tabular}{l|cccccc}
    \Xhline{1pt}
    Method 
    &  
    PCK
    &
    AUC
    &
    MPJPE
    & 
    PA-MPJPE
    & 
    PVE
    &
    Time[s]
    \\
    \hline
        SPIN~\cite{kolotouros2019spin} & 30.8 & 53.4 & 99.4 & 68.1 & -  & 0.31 \\
        VIBE~\cite{kocabas2020vibe} & \textbf{33.9} & \textbf{56.6} & \textbf{94.7} & 66.1 & \textbf{112.7} &  - \\
        CRMH~\cite{jiang2020coherent} & 25.8 & 51.6  & 105.3 & \textbf{62.3}  & 122.2 &  0.09 \\
     	BMP-Lite & 26.2 & 51.3 & 108.5 & 64.0 & 126.2 &  \textbf{0.038} \\
        BMP & 32.1 & 54.5 & 104.1 & 63.8 & 119.3 &  0.056 \\
    \Xhline{1pt}
    \end{tabular}
    \caption{\textbf{Results on 3DPW.} We use 3DPCK, AUC, MPJPE, PA-MPJPE and per-vertex error (PVE) as evaluation metrics.  }  \label{tab:3dpw}%
\end{table}

\begin{figure*}[!ht]
	\centering
    \includegraphics[width=0.24\textwidth]{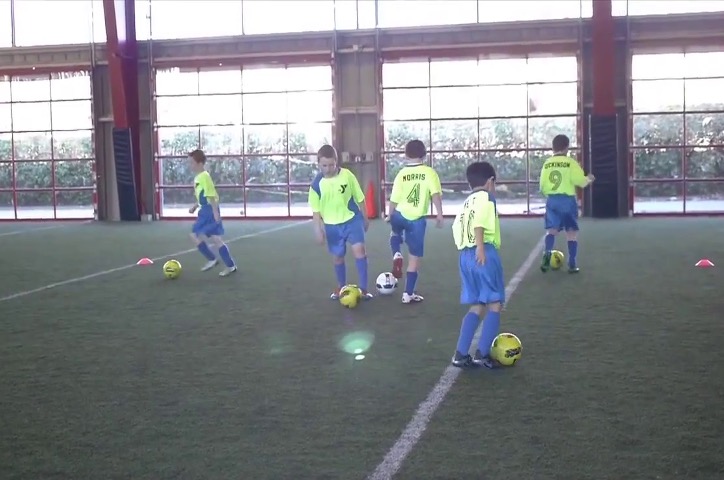}
    \includegraphics[width=0.24\textwidth]{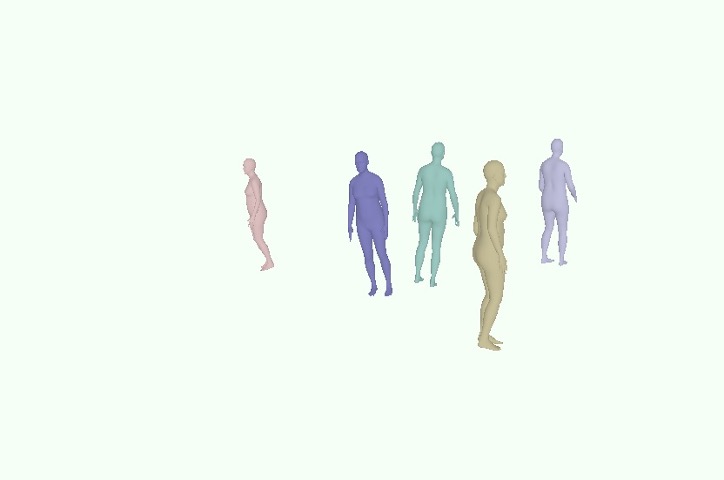}
    \includegraphics[width=0.24\textwidth]{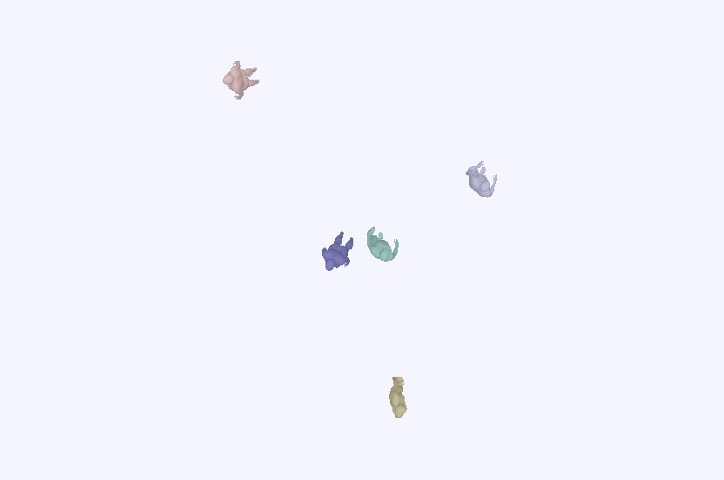}
    \includegraphics[width=0.24\textwidth]{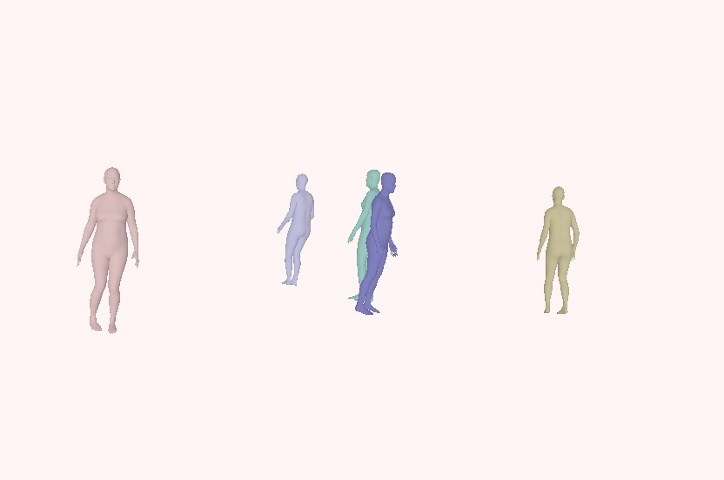}
    \\
    \includegraphics[width=0.24\textwidth]{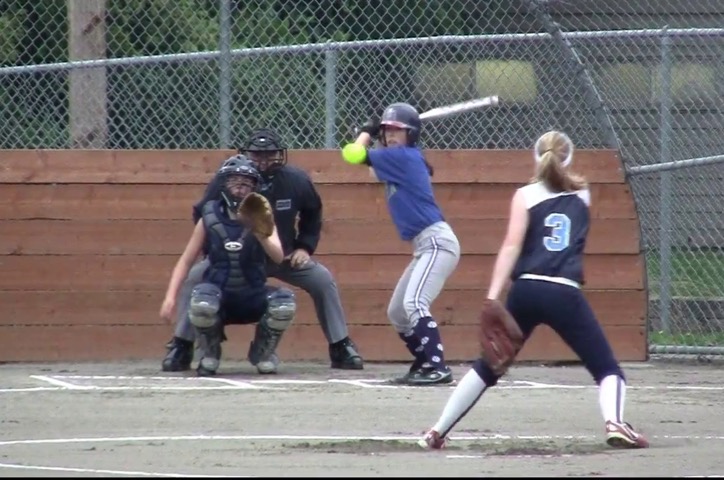}
    \includegraphics[width=0.24\textwidth]{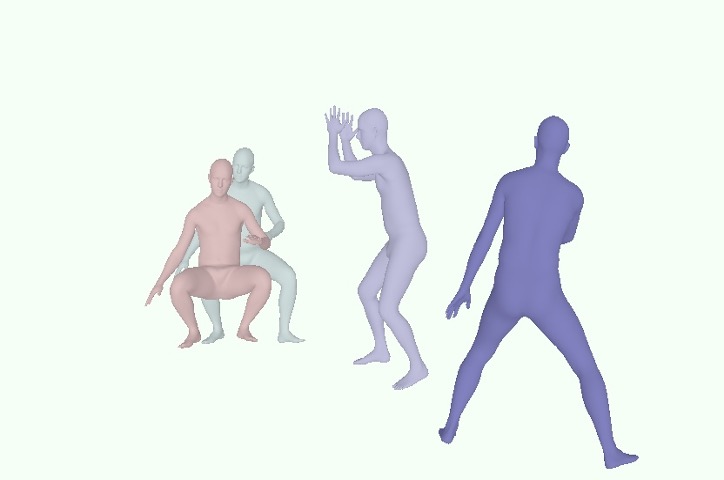}
    \includegraphics[width=0.24\textwidth]{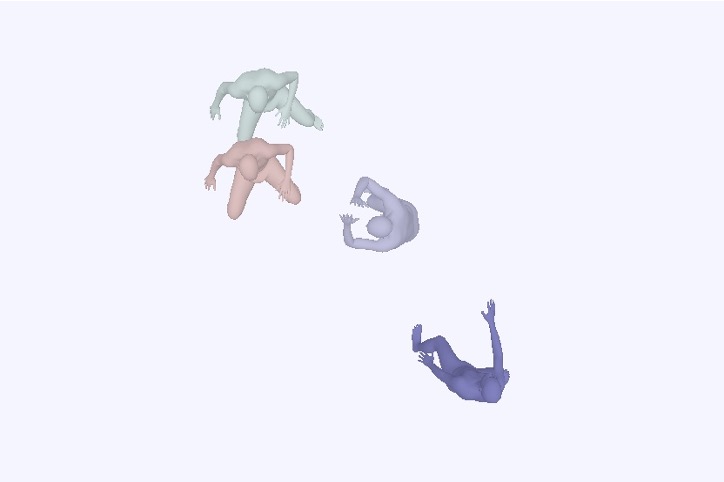}
    \includegraphics[width=0.24\textwidth]{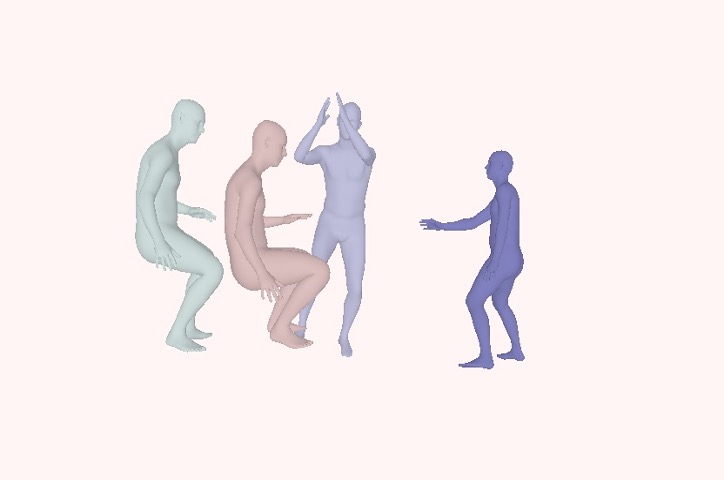}
    \\
    \includegraphics[width=0.24\textwidth]{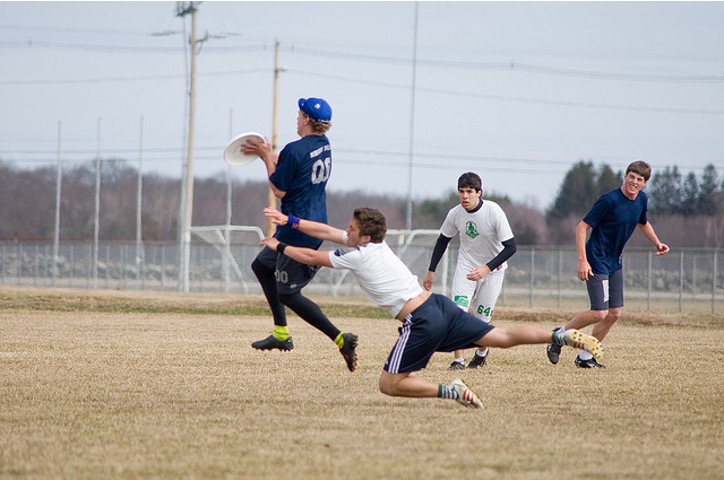}
    \includegraphics[width=0.24\textwidth]{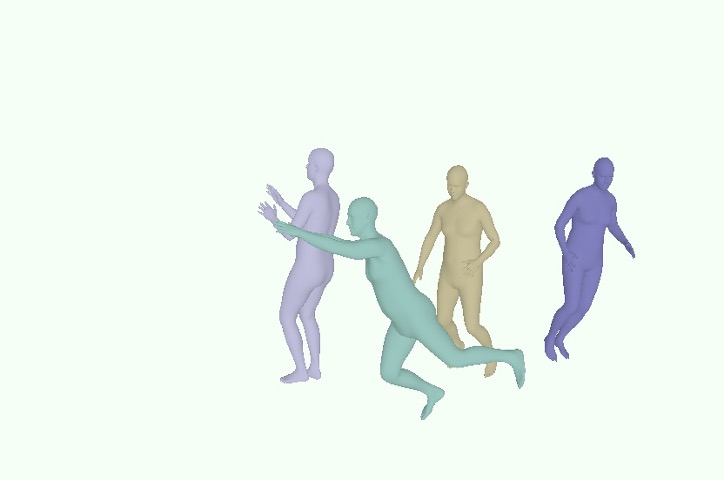}
    \includegraphics[width=0.24\textwidth]{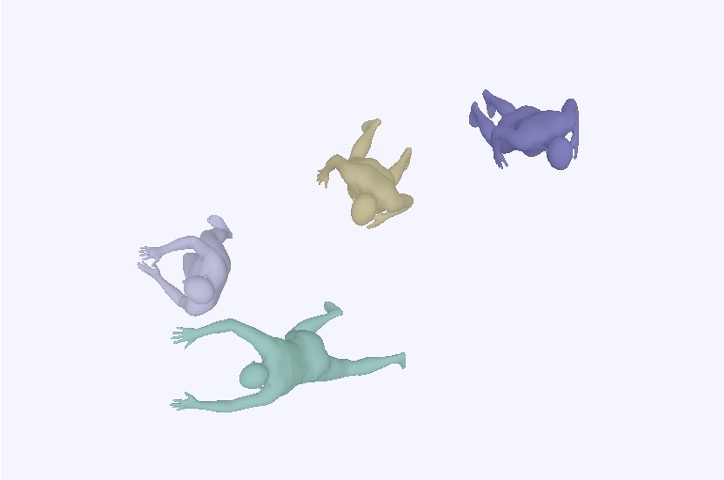}
    \includegraphics[width=0.24\textwidth]{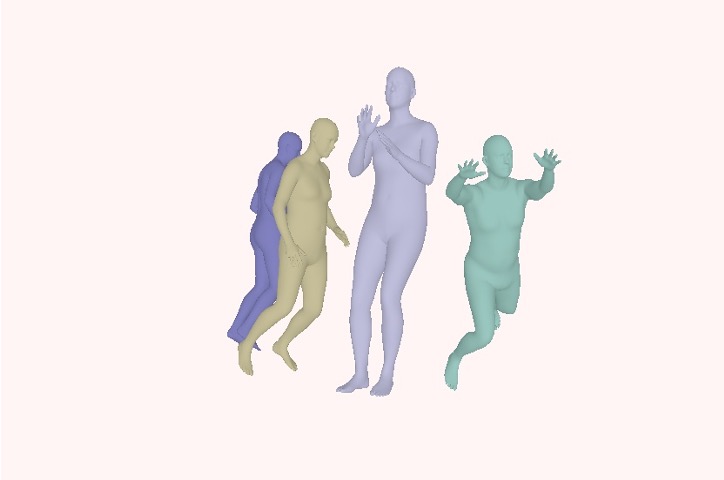}
    \\
	\caption{{\bf Qualitative results.}
    We visualize the reconstructions of our approach on PoseTrack (1st row), MPII (2nd row) and COCO (3rd row) from different viewpoints: front (green background), top (blue background) and side (red background), respectively. Best viewed in color.
    Please refer to supplementary for more qualitative results.
}
\label{fig:qualitative}
\vspace{-4mm}
\end{figure*}

\myparagraph{Qualitative results}
We visualize some body mesh reconstructions of BMP on the challenging PoseTrack, MPII and COCO datasets, as shown in Fig.~\ref{fig:qualitative}. It can be observed that BMP is robust to severe occlusion and crowded scenes and can reconstruct human bodies with correct depth ordering.

\subsection{Ablative studies} \label{sec:ablatives}

\begin{figure*}[!h]
	\centering
    \includegraphics[width=0.24\textwidth]{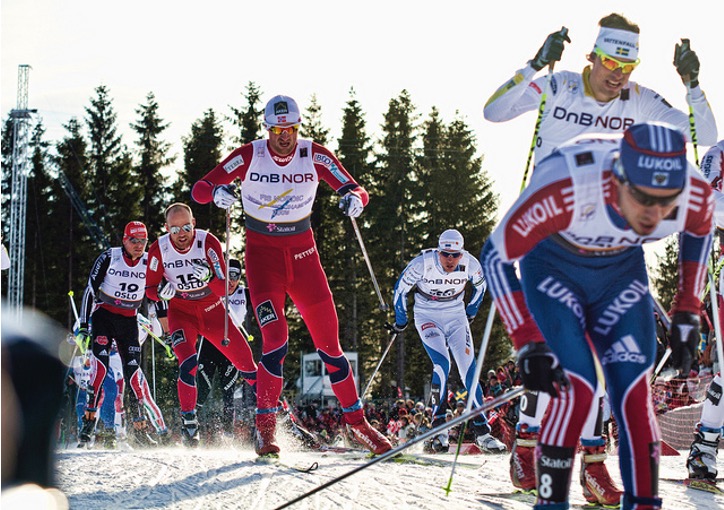}
    \includegraphics[width=0.24\textwidth]{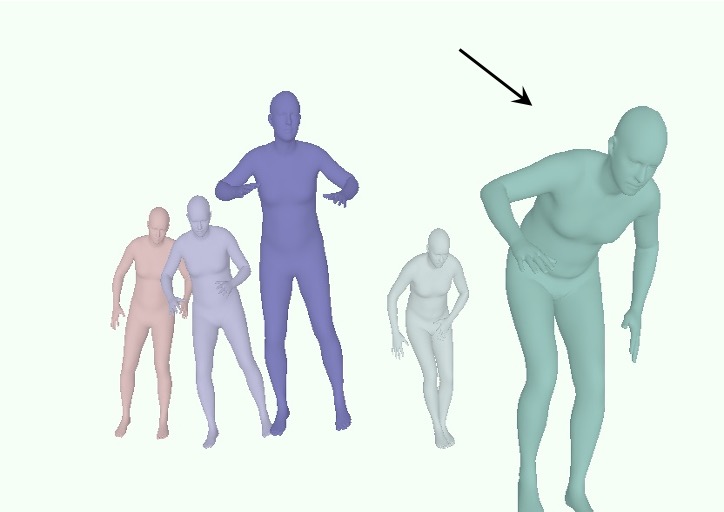}
    \includegraphics[width=0.12\textwidth]{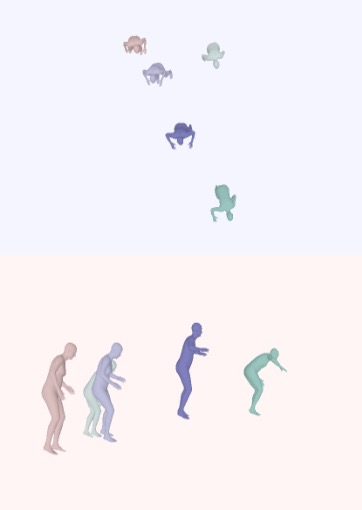}
    \includegraphics[width=0.24\textwidth]{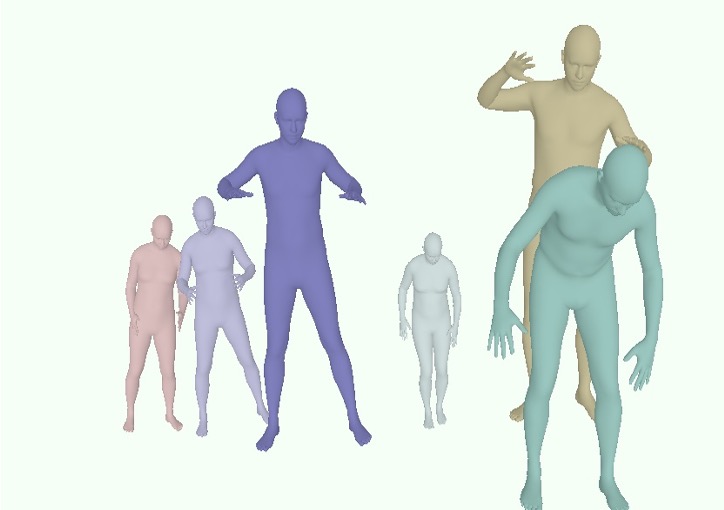}
    \includegraphics[width=0.12\textwidth]{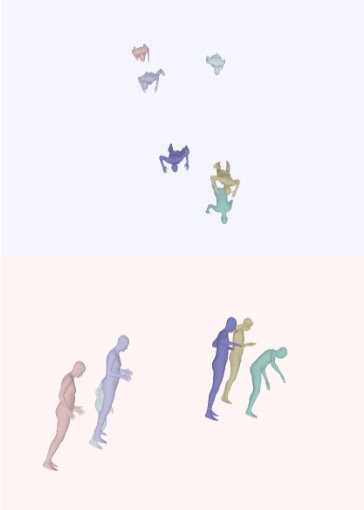}\\
    \includegraphics[width=0.24\textwidth]{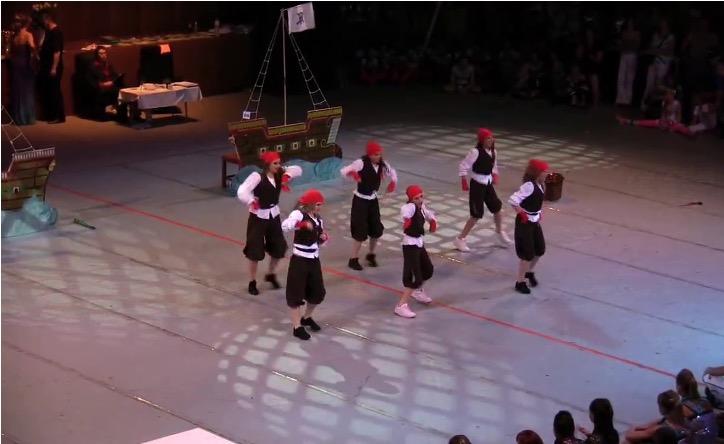}
    \includegraphics[width=0.24\textwidth]{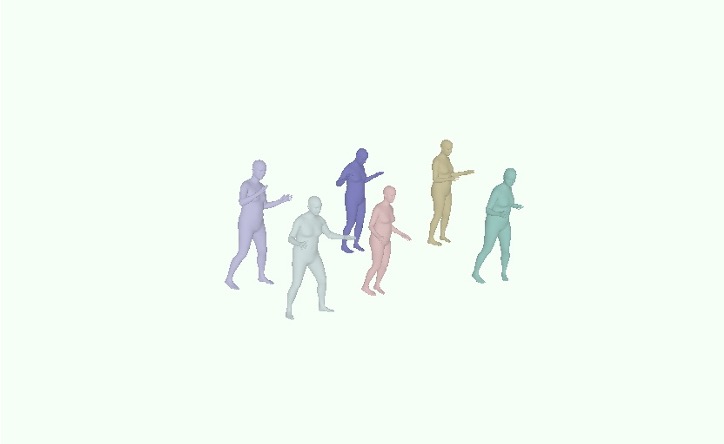}
    \includegraphics[width=0.12\textwidth]{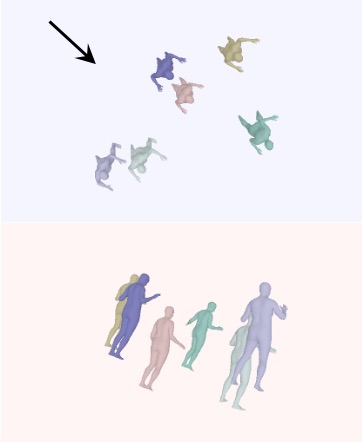}
    \includegraphics[width=0.24\textwidth]{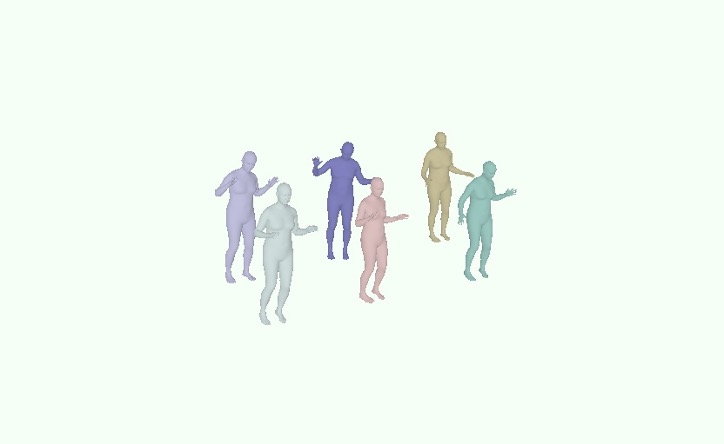}
    \includegraphics[width=0.12\textwidth]{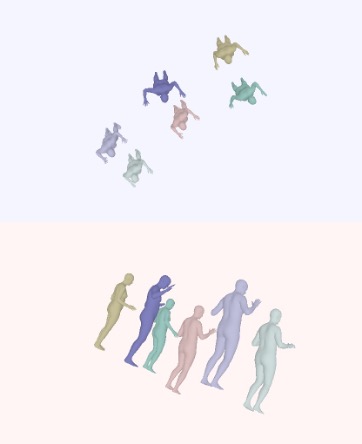}\\
  \vspace{-1mm}
   \text{\scriptsize Input image } \hspace{40mm} \text{\scriptsize Baseline} \hspace{56mm} \text{\scriptsize Ours}  \hspace{12mm} \text{\color{white} \scriptsize}
	\caption{{\bf Qualitative effect of proposed method.}
	Results of baseline 1 (BMP using 2D representation) (middle 1st row), baseline 2 (BMP trained w/o $L_{rank}$) (middle 2rd row) and BMP (right).  Errors are highlighted by black arrows. As expected, the proposed methods take effect on producing better results (\ie, robust to overlapping instances, more consistent depth ordering for estimated body meshes).
}
\label{fig:ablative}
\end{figure*}

We conduct ablation analysis on Panoptic, 3DPW and MuPoTS-3D datasets both qualitatively and quantitatively to justify our design choices. The qualitative analysis of the proposed method is illustrated in Fig.~\ref{fig:ablative}.

\myparagraph{Person instance representation}
We first evaluate the proposed 3D point-based representation for person instances. 
The main difference between the proposed representation and previous 2D spatial representation~\cite{nie2019spm,zhou2019objects,CenterHMR} is that we use an additional depth dimension to differentiate person instances in the discretized depth space through FPN.
We then compare BMP with a baseline model (\ie, BMP using 2D spatial representation).
For fair comparison, we aggregate features from all levels of FPN pyramid in the baseline model to obtain a single output for both instance localization and body mesh recovery. 
Specifically, we study three methods for the aggregation:
we resize all feature pyramids to 1/8 scale and then aggregate them by 1) element-wise addition (\textit{Baseline-Add}), 2) concatenation (\textit{Baseline-Concat}), or 3) adopting a convolutional layer after concatenating them (\textit{Baseline-Conv}).
Results are shown in Table~\ref{tab:ablation-deplayering}. 
We can see BMP improves upon the baseline models by a large margin on all datasets, proving its efficacy for body mesh recovery. 
Additionally, from Fig.~\ref{fig:ablative} (1st row), we observe BMP with the proposed representation is more robust in handling occluding instances, especially when the body centers of multiple instances fall at the same spatial grid coordinate, while the 2D representation would usually fail. 

\begin{table}[!h]
\setlength\tabcolsep{2.5mm}
\footnotesize
\centering
	\begin{tabular}{l|ccc}
		\Xhline{1pt}
		Method & Panoptic ($\downarrow$) & 3DPW ($\downarrow$) & MuPoTS-3D ($\uparrow$)  \\
		\hline 
		Baseline-Add & 159.1 & 120.4 & 68.03
		\\
		Baseline-Concat & 150.3 & 114.6 & 68.52
		\\
		Baseline-Conv & 145.6 & 110.8 & 69.34
		\\
		BMP & 135.4 & 104.1 & 73.83
		\\
		\Xhline{1pt}
	\end{tabular}
	\caption{\textbf{Ablation for person instance representation.} We report MPJPE for Panoptic and 3DPW, and 3DPCK for MuPoTS-3D.
	}
	\label{tab:ablation-deplayering}
\end{table}

\myparagraph{Ordinal depth loss}
To investigate whether the ordinal depth loss $L_{rank}$ can help produce more coherent results with correct depth ordering, we conduct experiments on the MuPoTS-3D dataset. 
Specifically, we evaluate the ordinal depth relations of all instance pairs in the scene and report the percentage of correctly estimated ordinal depth relations in Table~\ref{tab:ablation-ordering}.
The model trained with $L_{rank}$ significantly improves upon the baseline (BMP trained w/o $L_{rank}$)  (from 91.42\% to 94.50\%). Such improvements can also be observed from Fig.~\ref{fig:ablative} (2nd row).
Additionally, by comparing our method with Moon~\et~\cite{moon2019camera} and CRMH~\cite{jiang2020coherent}, we observe BMP achieves higher accuracy w.r.t.\ relative depth ordering than CRMH that only considers ordinal loss for overlapped pairs (94.50\% \vs 93.68\%).
This demonstrates  our full pair-wise ordinal loss can provide a more comprehensive supervision on the depth layout of the scene and thus train the model to give more coherent results.

\begin{table}[!ht]
\setlength\tabcolsep{1.5mm}
\footnotesize
\centering
	\begin{tabular}{c|c|c|c|c}
		\Xhline{1pt}
		Method & Moon~\cite{moon2019camera} & CRMH~\cite{jiang2020coherent} & BMP w/o $L_{rank}$ & BMP \\
		\hline 
		Accuracy & 90.85\% & 93.68\% & 91.42\% & 94.50\%
		\\
		\Xhline{1pt}
	\end{tabular}
	\caption{\textbf{Ablation for ordinal depth loss.} Relative depth ordering results on MuPoTS-3D are shown. We evaluate the ordinal depth relations of all instance pairs in the scene and report the percentage of correctly estimated ordinal depth relations. 
	}
	\label{tab:ablation-ordering}
\end{table}

\myparagraph{Keypoint-aware occlusion augmentation}
Finally, we study the impact of the proposed keypoint-aware occlusion augmentation strategy. 
We compare our BMP model with the models trained without occlusion augmentation (\textit{BMP-NoAug}) and trained using randomly Synthetic Occlusion~\cite{sarandi2018robust} (\textit{BMP-RandOcc}) in Table~\ref{tab:ablation-aug}. 
We can see BMP outperforms both of them  by a large margin on all datasets. 
Notably, it respectively brings 9.1\% and 17.3\% improvements over BMP-NoAug on Panoptic and 3DPW datasets, which feature crowded scenes with severe overlap and partial observation.
In contrast, the random augmentation hurts model performance on MuPoTS-3D (71.71 \vs 70.78).
This verifies that our proposed occlusion augmentation can force the model to focus on body structure and thus improve its robustness to occlusion.

\begin{table}[!ht]
\setlength\tabcolsep{2mm}

\footnotesize
\centering
	\begin{tabular}{l|ccc}
		\Xhline{1pt}
		Method & Panoptic ($\downarrow$) & 3DPW ($\downarrow$) & MuPoTS-3D ($\uparrow$)  \\
		\hline 
		BMP-NoAug & 148.9 & 125.9 & 71.71
		\\
		BMP-RandOcc & 144.6 & 110.3 & 70.78
		\\
		BMP & 135.4 & 104.1 & 73.83
		\\
		\Xhline{1pt}
	\end{tabular}
	\caption{\textbf{Ablation for occlusion augmentation.} We use MPJPE for the first two, and 3DPCK for the last one datasets as metrics.
	}
	\label{tab:ablation-aug}
\end{table}

\section{Conclusions}

In this work, we present the first single-stage model, Body Meshes as Points (BMP), for multi-person body mesh recovery. BMP introduces a new representation method to enable such a compact pipeline: each person instance is represented as a point in the spatial-depth space which is associated with a parameterized body mesh. With such a representation, BMP can fully exploit shared features and perform person localization and body mesh recovery simultaneously. BMP significantly improves upon conventional two-stage paradigms, and offers outstanding efficiency and accuracy, as validated by extensive experiments on multiple benchmarks. Besides, BMP develops several new techniques to further improve the coherence and robustness of recovered body meshes, which are of broad interest for other applications like human pose estimation and detection. In future, we will explore how to make the model more compact and further improve its efficiency, as well as extend to inter-person interactions modeling.

\myparagraph{Acknowledgement}
This research was partially supported by AISG-100E-2019-035, MOE2017-T2-2-151, NUS\_ECRA\_FY17\_P08 and CRP20-2017-0006.

{\small
\bibliographystyle{ieee_fullname}
\bibliography{egbib}

\begin{thebibliography}{10}\itemsep=-1pt

\bibitem{Andriluka_2018_CVPR}
Mykhaylo Andriluka, Umar Iqbal, Eldar Insafutdinov, Leonid Pishchulin, Anton
  Milan, Juergen Gall, and Bernt Schiele.
\newblock Posetrack: A benchmark for human pose estimation and tracking.
\newblock In {\em {CVPR}}, 2018.

\bibitem{andriluka20142d}
Mykhaylo Andriluka, Leonid Pishchulin, Peter Gehler, and Bernt Schiele.
\newblock 2d human pose estimation: New benchmark and state of the art
  analysis.
\newblock In {\em {CVPR}}, 2014.

\bibitem{arnab2019exploiting}
Anurag Arnab, Carl Doersch, and Andrew Zisserman.
\newblock Exploiting temporal context for 3d human pose estimation in the wild.
\newblock In {\em {CVPR}}, 2019.

\bibitem{benzine2020pandanet}
Abdallah Benzine, Florian Chabot, Bertrand Luvison, Quoc~Cuong Pham, and
  Catherine Achard.
\newblock Pandanet: Anchor-based single-shot multi-person 3d pose estimation.
\newblock In {\em {CVPR}}, 2020.

\bibitem{bogo2016keep}
Federica Bogo, Angjoo Kanazawa, Christoph Lassner, Peter Gehler, Javier Romero,
  and Michael~J Black.
\newblock Keep it smpl: Automatic estimation of 3d human pose and shape from a
  single image.
\newblock In {\em {ECCV}}, 2016.

\bibitem{cao2018openpose}
Zhe Cao, Gines Hidalgo, Tomas Simon, Shih-En Wei, and Yaser Sheikh.
\newblock Openpose: realtime multi-person 2d pose estimation using part
  affinity fields.
\newblock {\em arXiv}, 2018.

\bibitem{cao2017realtime}
Zhe Cao, Tomas Simon, Shih-En Wei, and Yaser Sheikh.
\newblock Realtime multi-person 2d pose estimation using part affinity fields.
\newblock In {\em {CVPR}}, 2017.

\bibitem{mmdetection}
Kai Chen, Jiaqi Wang, Jiangmiao Pang, Yuhang Cao, Yu Xiong, Xiaoxiao Li,
  Shuyang Sun, Wansen Feng, Ziwei Liu, Jiarui Xu, Zheng Zhang, Dazhi Cheng,
  Chenchen Zhu, Tianheng Cheng, Qijie Zhao, Buyu Li, Xin Lu, Rui Zhu, Yue Wu,
  Jifeng Dai, Jingdong Wang, Jianping Shi, Wanli Ouyang, Chen~Change Loy, and
  Dahua Lin.
\newblock {MMDetection}: Open mmlab detection toolbox and benchmark.
\newblock {\em arXiv}, 2019.

\bibitem{dabral2018learning}
Rishabh Dabral, Anurag Mundhada, Uday Kusupati, Safeer Afaque, Abhishek Sharma,
  and Arjun Jain.
\newblock Learning 3d human pose from structure and motion.
\newblock In {\em {ECCV}}, 2018.

\bibitem{duan2019centernet}
Kaiwen Duan, Song Bai, Lingxi Xie, Honggang Qi, Qingming Huang, and Qi Tian.
\newblock Centernet: Keypoint triplets for object detection.
\newblock In {\em {ICCV}}, 2019.

\bibitem{everingham2011pascal}
Mark Everingham and John Winn.
\newblock The pascal visual object classes challenge 2012 (voc2012) development
  kit.
\newblock {\em {IEEE Trans. on Pattern Analysis and Machine Intelligence}}, 8,
  2011.

\bibitem{fabbri2020compressed}
Matteo Fabbri, Fabio Lanzi, Simone Calderara, Stefano Alletto, and Rita
  Cucchiara.
\newblock Compressed volumetric heatmaps for multi-person 3d pose estimation.
\newblock In {\em {CVPR}}, 2020.

\bibitem{gabeur2019moulding}
Valentin Gabeur, Jean-S{\'e}bastien Franco, Xavier Martin, Cordelia Schmid, and
  Gregory Rogez.
\newblock Moulding humans: Non-parametric 3d human shape estimation from single
  images.
\newblock In {\em {CVPR}}, 2019.

\bibitem{girdhar2018detect}
Rohit Girdhar, Georgia Gkioxari, Lorenzo Torresani, Manohar Paluri, and Du
  Tran.
\newblock Detect-and-track: Efficient pose estimation in videos.
\newblock In {\em {CVPR}}, 2018.

\bibitem{gong2021poseaug}
Kehong Gong, Jianfeng Zhang, and Jiashi Feng.
\newblock Poseaug: A differentiable pose augmentation framework for 3d human
  pose estimation.
\newblock In {\em {CVPR}}, 2021.

\bibitem{guler2019holopose}
Riza~Alp Guler and Iasonas Kokkinos.
\newblock Holopose: Holistic 3d human reconstruction in-the-wild.
\newblock In {\em {CVPR}}, 2019.

\bibitem{guo2020pi}
Wen Guo, Enric Corona, Francesc Moreno-Noguer, and Xavier Alameda-Pineda.
\newblock Pi-net: Pose interacting network for multi-person monocular 3d pose
  estimation.
\newblock {\em arXiv}, 2020.

\bibitem{He_2016_CVPR}
Kaiming He, Xiangyu Zhang, Shaoqing Ren, and Jian Sun.
\newblock Deep residual learning for image recognition.
\newblock In {\em {CVPR}}, 2016.

\bibitem{ionescu2014human3}
Catalin Ionescu, Dragos Papava, Vlad Olaru, and Cristian Sminchisescu.
\newblock Human3. 6m: Large scale datasets and predictive methods for 3d human
  sensing in natural environments.
\newblock {\em {IEEE Trans. on Pattern Analysis and Machine Intelligence}},
  36(7):1325--1339, 2014.

\bibitem{jiang2020coherent}
Wen Jiang, Nikos Kolotouros, Georgios Pavlakos, Xiaowei Zhou, and Kostas
  Daniilidis.
\newblock Coherent reconstruction of multiple humans from a single image.
\newblock In {\em {CVPR}}, 2020.

\bibitem{johnson2010clustered}
Sam Johnson and Mark Everingham.
\newblock Clustered pose and nonlinear appearance models for human pose
  estimation.
\newblock In {\em {BMVC}}, 2010.

\bibitem{Johnson11}
Sam Johnson and Mark Everingham.
\newblock Learning effective human pose estimation from inaccurate annotation.
\newblock In {\em {CVPR}}, 2011.

\bibitem{joo2015panoptic}
Hanbyul Joo, Hao Liu, Lei Tan, Lin Gui, Bart Nabbe, Iain Matthews, Takeo
  Kanade, Shohei Nobuhara, and Yaser Sheikh.
\newblock Panoptic studio: A massively multiview system for social motion
  capture.
\newblock In {\em {ICCV}}, 2015.

\bibitem{hmrKanazawa17}
Angjoo Kanazawa, Michael~J. Black, David~W. Jacobs, and Jitendra Malik.
\newblock End-to-end recovery of human shape and pose.
\newblock In {\em {CVPR}}, 2018.

\bibitem{kocabas2020vibe}
Muhammed Kocabas, Nikos Athanasiou, and Michael~J Black.
\newblock Vibe: Video inference for human body pose and shape estimation.
\newblock In {\em {CVPR}}, 2020.

\bibitem{kolotouros2019spin}
Nikos Kolotouros, Georgios Pavlakos, Michael~J Black, and Kostas Daniilidis.
\newblock Learning to reconstruct 3d human pose and shape via model-fitting in
  the loop.
\newblock In {\em {ICCV}}, 2019.

\bibitem{kolotouros2019convolutional}
Nikos Kolotouros, Georgios Pavlakos, and Kostas Daniilidis.
\newblock Convolutional mesh regression for single-image human shape
  reconstruction.
\newblock In {\em {CVPR}}, 2019.

\bibitem{lassner2017unite}
Christoph Lassner, Javier Romero, Martin Kiefel, Federica Bogo, Michael~J
  Black, and Peter~V Gehler.
\newblock Unite the people: Closing the loop between 3d and 2d human
  representations.
\newblock In {\em {CVPR}}, 2017.

\bibitem{li2020hmor}
Jiefeng Li, Can Wang, Wentao Liu, Chen Qian, and Cewu Lu.
\newblock Hmor: Hierarchical multi-person ordinal relations for monocular
  multi-person 3d pose estimation.
\newblock In {\em {ECCV}}, 2020.

\bibitem{lin2017feature}
Tsung-Yi Lin, Piotr Doll{\'a}r, Ross Girshick, Kaiming He, Bharath Hariharan,
  and Serge Belongie.
\newblock Feature pyramid networks for object detection.
\newblock In {\em {CVPR}}, 2017.

\bibitem{lin2017focal}
Tsung-Yi Lin, Priya Goyal, Ross Girshick, Kaiming He, and Piotr Doll{\'a}r.
\newblock Focal loss for dense object detection.
\newblock In {\em {ICCV}}, 2017.

\bibitem{lin2014microsoft}
Tsung-Yi Lin, Michael Maire, Serge Belongie, James Hays, Pietro Perona, Deva
  Ramanan, Piotr Doll{\'a}r, and C~Lawrence Zitnick.
\newblock Microsoft coco: Common objects in context.
\newblock In {\em {ECCV}}, 2014.

\bibitem{liu2019variance}
Liyuan Liu, Haoming Jiang, Pengcheng He, Weizhu Chen, Xiaodong Liu, Jianfeng
  Gao, and Jiawei Han.
\newblock On the variance of the adaptive learning rate and beyond.
\newblock {\em arXiv}, 2019.

\bibitem{liu2018intriguing}
Rosanne Liu, Joel Lehman, Piero Molino, Felipe~Petroski Such, Eric Frank, Alex
  Sergeev, and Jason Yosinski.
\newblock An intriguing failing of convolutional neural networks and the
  coordconv solution.
\newblock In {\em {NeurIPS}}, 2018.

\bibitem{liu2019liquid}
Wen Liu, Zhixin Piao, Jie Min, Wenhan Luo, Lin Ma, and Shenghua Gao.
\newblock Liquid warping gan: A unified framework for human motion imitation,
  appearance transfer and novel view synthesis.
\newblock In {\em {ICCV}}, 2019.

\bibitem{loper2015smpl}
Matthew Loper, Naureen Mahmood, Javier Romero, Gerard Pons-Moll, and Michael~J
  Black.
\newblock Smpl: A skinned multi-person linear model.
\newblock {\em ACM transactions on graphics (TOG)}, 34(6):1--16, 2015.

\bibitem{martinez2017simple}
Julieta Martinez, Rayat Hossain, Javier Romero, and James~J Little.
\newblock A simple yet effective baseline for 3d human pose estimation.
\newblock In {\em {ICCV}}, 2017.

\bibitem{singleshotmultiperson2018}
Dushyant Mehta, Oleksandr Sotnychenko, Franziska Mueller, Weipeng Xu, Srinath
  Sridhar, Gerard Pons-Moll, and Christian Theobalt.
\newblock Single-shot multi-person 3d pose estimation from monocular rgb.
\newblock In {\em 3DV}, 2018.

\bibitem{mehta2018single}
Dushyant Mehta, Oleksandr Sotnychenko, Franziska Mueller, Weipeng Xu, Srinath
  Sridhar, Gerard Pons-Moll, and Christian Theobalt.
\newblock Single-shot multi-person 3d pose estimation from monocular rgb.
\newblock In {\em {3DV}}, 2018.

\bibitem{mehta2017vnect}
Dushyant Mehta, Srinath Sridhar, Oleksandr Sotnychenko, Helge Rhodin, Mohammad
  Shafiei, Hans-Peter Seidel, Weipeng Xu, Dan Casas, and Christian Theobalt.
\newblock Vnect: Real-time 3d human pose estimation with a single rgb camera.
\newblock {\em {ACM Trans. on Graphics}}, 36(4):44, 2017.

\bibitem{mir2020learning}
Aymen Mir, Thiemo Alldieck, and Gerard Pons-Moll.
\newblock Learning to transfer texture from clothing images to 3d humans.
\newblock In {\em {CVPR}}, 2020.

\bibitem{moon2019camera}
Gyeongsik Moon, Ju~Yong Chang, and Kyoung~Mu Lee.
\newblock Camera distance-aware top-down approach for 3d multi-person pose
  estimation from a single rgb image.
\newblock In {\em {ICCV}}, 2019.

\bibitem{nie2019spm}
Xuecheng Nie, Jianfeng Zhang, Shuicheng Yan, and Jiashi Feng.
\newblock Single-stage multi-person pose machines.
\newblock In {\em {ICCV}}, 2019.

\bibitem{omran2018neural}
Mohamed Omran, Christoph Lassner, Gerard Pons-Moll, Peter Gehler, and Bernt
  Schiele.
\newblock Neural body fitting: Unifying deep learning and model based human
  pose and shape estimation.
\newblock In {\em {3DV}}, 2018.

\bibitem{paszke2017automatic}
Adam Paszke, Sam Gross, Soumith Chintala, Gregory Chanan, Edward Yang, Zachary
  DeVito, Zeming Lin, Alban Desmaison, Luca Antiga, and Adam Lerer.
\newblock Automatic differentiation in pytorch.
\newblock In {\em NeurIPSw}, 2017.

\bibitem{pavlakos2019expressive}
Georgios Pavlakos, Vasileios Choutas, Nima Ghorbani, Timo Bolkart, Ahmed~AA
  Osman, Dimitrios Tzionas, and Michael~J Black.
\newblock Expressive body capture: 3d hands, face, and body from a single
  image.
\newblock In {\em {CVPR}}, 2019.

\bibitem{pavlakos2018ordinal}
Georgios Pavlakos, Xiaowei Zhou, and Kostas Daniilidis.
\newblock Ordinal depth supervision for 3d human pose estimation.
\newblock In {\em {CVPR}}, 2018.

\bibitem{pavlakos2018learning}
Georgios Pavlakos, Luyang Zhu, Xiaowei Zhou, and Kostas Daniilidis.
\newblock Learning to estimate 3d human pose and shape from a single color
  image.
\newblock In {\em {CVPR}}, 2018.

\bibitem{popa2017deep}
Alin-Ionut Popa, Mihai Zanfir, and Cristian Sminchisescu.
\newblock Deep multitask architecture for integrated 2d and 3d human sensing.
\newblock In {\em {CVPR}}, 2017.

\bibitem{redmon2018yolov3}
Joseph Redmon and Ali Farhadi.
\newblock Yolov3: An incremental improvement.
\newblock {\em arXiv}, 2018.

\bibitem{ren2015faster}
Shaoqing Ren, Kaiming He, Ross Girshick, and Jian Sun.
\newblock Faster r-cnn: Towards real-time object detection with region proposal
  networks.
\newblock In {\em {NeurIPS}}, 2015.

\bibitem{rockwell2020full}
Chris Rockwell and David~F Fouhey.
\newblock Full-body awareness from partial observations.
\newblock In {\em {ECCV}}, 2020.

\bibitem{rogez2017lcr}
Gregory Rogez, Philippe Weinzaepfel, and Cordelia Schmid.
\newblock Lcr-net: Localization-classification-regression for human pose.
\newblock In {\em {CVPR}}, 2017.

\bibitem{rogez2019lcr}
Gregory Rogez, Philippe Weinzaepfel, and Cordelia Schmid.
\newblock Lcr-net++: Multi-person 2d and 3d pose detection in natural images.
\newblock {\em {IEEE Trans. on Pattern Analysis and Machine Intelligence}},
  42(5):1146--1161, 2019.

\bibitem{sarandi2018robust}
Istv{\'a}n S{\'a}r{\'a}ndi, Timm Linder, Kai~O Arras, and Bastian Leibe.
\newblock How robust is 3d human pose estimation to occlusion?
\newblock In {\em {ICRAw}}, 2018.

\bibitem{smith2019facsimile}
David Smith, Matthew Loper, Xiaochen Hu, Paris Mavroidis, and Javier Romero.
\newblock Facsimile: Fast and accurate scans from an image in less than a
  second.
\newblock In {\em {ICCV}}, 2019.

\bibitem{su2019multi}
Kai Su, Dongdong Yu, Zhenqi Xu, Xin Geng, and Changhu Wang.
\newblock Multi-person pose estimation with enhanced channel-wise and spatial
  information.
\newblock In {\em {CVPR}}, 2019.

\bibitem{sun2018integral}
Xiao Sun, Bin Xiao, Fangyin Wei, Shuang Liang, and Yichen Wei.
\newblock Integral human pose regression.
\newblock In {\em {ECCV}}, 2018.

\bibitem{tian2019fcos}
Zhi Tian, Chunhua Shen, Hao Chen, and Tong He.
\newblock Fcos: Fully convolutional one-stage object detection.
\newblock In {\em {ICCV}}, 2019.

\bibitem{tome2017lifting}
Denis Tome, Chris Russell, and Lourdes Agapito.
\newblock Lifting from the deep: Convolutional 3d pose estimation from a single
  image.
\newblock In {\em {CVPR}}, 2017.

\bibitem{tung2017self}
Hsiao-Yu Tung, Hsiao-Wei Tung, Ersin Yumer, and Katerina Fragkiadaki.
\newblock Self-supervised learning of motion capture.
\newblock In {\em {NeurIPS}}, 2017.

\bibitem{varol2018bodynet}
Gul Varol, Duygu Ceylan, Bryan Russell, Jimei Yang, Ersin Yumer, Ivan Laptev,
  and Cordelia Schmid.
\newblock Bodynet: Volumetric inference of 3d human body shapes.
\newblock In {\em {ECCV}}, 2018.

\bibitem{varol2019synthetic}
G{\"u}l Varol, Ivan Laptev, Cordelia Schmid, and Andrew Zisserman.
\newblock Synthetic humans for action recognition from unseen viewpoints.
\newblock {\em arXiv}, 2019.

\bibitem{vonMarcard2018}
Timo von Marcard, Roberto Henschel, Michael Black, Bodo Rosenhahn, and Gerard
  Pons-Moll.
\newblock Recovering accurate 3d human pose in the wild using imus and a moving
  camera.
\newblock In {\em {ECCV}}, 2018.

\bibitem{wang2019solo}
Xinlong Wang, Tao Kong, Chunhua Shen, Yuning Jiang, and Lei Li.
\newblock Solo: Segmenting objects by locations.
\newblock In {\em {ECCV}}, 2019.

\bibitem{weinzaepfel2020dope}
Philippe Weinzaepfel, Romain Br{\'e}gier, Hadrien Combaluzier, Vincent Leroy,
  and Gr{\'e}gory Rogez.
\newblock Dope: Distillation of part experts for whole-body 3d pose estimation
  in the wild.
\newblock In {\em {ECCV}}, 2020.

\bibitem{wu2018group}
Yuxin Wu and Kaiming He.
\newblock Group normalization.
\newblock In {\em {ECCV}}, 2018.

\bibitem{CenterHMR}
Sun Yu, Bao Qian, Liu Wu, Fu Yili, and Mei Tao.
\newblock Centerhmr: a bottom-up single-shot method for multi-person 3d mesh
  recovery from a single image.
\newblock {\em arXiv}, 2020.

\bibitem{zanfir2018monocular}
Andrei Zanfir, Elisabeta Marinoiu, and Cristian Sminchisescu.
\newblock Monocular 3d pose and shape estimation of multiple people in natural
  scenes-the importance of multiple scene constraints.
\newblock In {\em {CVPR}}, 2018.

\bibitem{zanfir2018deep}
Andrei Zanfir, Elisabeta Marinoiu, Mihai Zanfir, Alin-Ionut Popa, and Cristian
  Sminchisescu.
\newblock Deep network for the integrated 3d sensing of multiple people in
  natural images.
\newblock In {\em {NeurIPS}}, 2018.

\bibitem{zhang2020inference}
Jianfeng Zhang, Xuecheng Nie, and Jiashi Feng.
\newblock Inference stage optimization for cross-scenario 3d human pose
  estimation.
\newblock In {\em {NeurIPS}}, 2020.

\bibitem{zhang2020object}
Tianshu Zhang, Buzhen Huang, and Yangang Wang.
\newblock Object-occluded human shape and pose estimation from a single color
  image.
\newblock In {\em {CVPR}}, 2020.

\bibitem{zhen2020smap}
Jianan Zhen, Qi Fang, Jiaming Sun, Wentao Liu, Wei Jiang, Hujun Bao, and
  Xiaowei Zhou.
\newblock Smap: Single-shot multi-person absolute 3d pose estimation.
\newblock In {\em {ECCV}}, 2020.

\bibitem{zhou2017towards}
Xingyi Zhou, Qixing Huang, Xiao Sun, Xiangyang Xue, and Yichen Wei.
\newblock Towards 3d human pose estimation in the wild: a weakly-supervised
  approach.
\newblock In {\em {ICCV}}, 2017.

\bibitem{zhou2019objects}
Xingyi Zhou, Dequan Wang, and Philipp Kr{\"a}henb{\"u}hl.
\newblock Objects as points.
\newblock {\em arXiv}, 2019.

\bibitem{zhou2019continuity}
Yi Zhou, Connelly Barnes, Jingwan Lu, Jimei Yang, and Hao Li.
\newblock On the continuity of rotation representations in neural networks.
\newblock In {\em {CVPR}}, 2019.

\end{thebibliography}
}

\end{document}